# AN INFORMATION EXTRACTION APPROACH TO PRESCREEN HEART FAILURE PATIENTS FOR CLINICAL TRIALS

ABHISHEK KALYAN ADUPA[1], RAVI PRAKASH GARG[1], JESSICA CORONA-COX[2], SANJIV J. SHAH[2], SIDDHARTHA JONNALAGADDA[1]

[1]*Division of Health and Biomedical Informatics, Department of Preventive Medicine, Northwestern University Feinberg School of Medicine, Chicago, IL 60611, USA*
*Email: sid@northwestern.edu*

[2]*Division of Cardiology, Department of Medicine, Northwestern University Feinberg School of Medicine, Chicago, IL 60611, USA*

To reduce the large amount of time spent screening, identifying, and recruiting patients into clinical trials, we need prescreening systems that are able to automate the data extraction and decision-making tasks that are typically relegated to clinical research study coordinators. However, a major obstacle is the vast amount of patient data available as unstructured free-form text in electronic health records. Here we propose an information extraction-based approach that first automatically converts unstructured text into a structured form. The structured data are then compared against a list of eligibility criteria using a rule-based system to determine which patients qualify for enrollment in a heart failure clinical trial. We show that we can achieve highly accurate results, with recall and precision values of 0.95 and 0.86, respectively. Our system allowed us to significantly reduce the time needed for prescreening patients from a few weeks to a few minutes. Our open-source information extraction modules are available for researchers and could be tested and validated in other cardiovascular trials. An approach such as the one we demonstrate here may decrease costs and expedite clinical trials, and could enhance the reproducibility of trials across institutions and populations.

## 1. Introduction

The creation and acceptance of electronic health records (EHRs) has ignited widespread interest in the use of clinical data for secondary purposes and research [1]. One such application that can greatly benefit from an EHR-based approach is clinical trial screening and recruitment. Clinical trial screening is a process that helps medical practitioners and researchers determine whether a particular patient is suitable for trial based on certain eligibility criteria. The eligibility criteria are generally divided into two parts: inclusion criteria and exclusion criteria. Inclusion criteria are characteristics that the prospective subjects must have if they are to be included in the study, while exclusion criteria are those characteristics that disqualify prospective subjects from inclusion in the study.

In general, screening for clinical trial recruitment is done manually. Clinicians and study coordinators go through each of the eligibility criteria, determine data elements relevant to the clinical trial, extract the data elements from structured and unstructured EHR of each patient, and match the data elements with the eligibility criteria to decide whether the patient qualifies for the trial. Not only this process is slow, it is also prone to errors. It typically takes approximately 15 to 20 minutes for a study coordinator to examine each patient's data. Because of the subjectivity involved in human decision-making, domain knowledge, which patients are considered for initial search and other factors [2], there is always a possibility of type-1 and type-2 errors in the prescreening process and biases in the overall recruitment. Furthermore, clinicians and study coordinators typically rely on patients identified in their own specialty clinics or in certain defined patient care settings, thereby missing out on the advantage of screening an entire healthcare system.

We hypothesize that an automated process for prescreening would be quicker and serve as an independent judge of inclusion/exclusion criteria free of human bias. If the prescreening algorithm also has a high recall (sensitivity), it would potentially reduce recruitment bias because it would be possible to consider patients from a larger pool. Thus, an algorithm that can prescreen eligible patients efficiently could provide a proficient and robust approach to clinical trial recruitment. Therefore, we sought to develop a high recall (sensitivity) prescreening algorithm for recruiting patients into a multicenter, randomized, double-blind, parallel group, active-controlled study to evaluate the efficacy and safety of LCZ696 compared to valsartan, on morbidity and mortality in heart failure patients with preserved ejection fraction (PARAGON). Our approach involves development of information extraction modules that can be reused not only for other EHRs but also for other trials using similar data elements.

## 2. Background

Heart failure (HF) occurs when the heart muscle is no longer able to meet the demands of the body either due to reduced cardiac output or increased ventricular filling pressures. It is one of the most common reasons for hospital admissions among those aged 65 years and older. In 2010 alone, HF affected 6.6 million Americans at a cost of $34.4 billion [3, 4]. Many clinical trials have been undertaken to find efficient solutions to the condition. However, it has been found that 86% of all clinical trials are delayed in patient recruitment from 1 to 6 months, and 13% are delayed by longer than 6 months [2]. A major cause of delay in HF clinical trials is the inability to efficiently screen for and identify eligible patients. An automated system is therefore needed to accelerate the process of prescreening patients for clinical trials.

The surge of the use of EHRs in the United States has created abundant opportunities for clinical and translational research. As Friedman et al noted, the extensive use of clinical data provides great potential to transform our healthcare system into a "Self-learning Health System" [5, 6]. In addition to its primary purpose of providing improved clinical practice, the use of EHRs offers means for the identification of participants who satisfy predefined criteria. This can be used for a variety of applications, including clinical trial recruitment, outcome prediction, survival analysis, and other retrospective studies [7-10].

EHRs contain patient data in both structured and unstructured formats. The structured data generally encompass a patient's demographic data, physical characteristics (e.g., body mass index [BMI], blood pressure), laboratory data, and diagnoses. Structured data are not only the best representation of knowledge but also easier to process. However, there is a vast amount of medical knowledge that is locked in the unstructured format. The unstructured data are typically text clinical narratives present in progress notes, imaging reports (e.g., echocardiographic reports), and discharge summaries, for example. Thus, a module that can automatically and efficiently extract information from unstructured clinical text and convert it into a structured form is needed.

The syndromic nature of HF presents unique challenges for the identification of patients from EHR data for research [11]. HF with preserved ejection fraction, in which the global pumping function of the heart is normal, is particularly challenging to identify during prescreening activities. The presence of large amounts of unstructured data in patient medical reports aggravates the challenges. Previous studies have shown that clinicians often prefer free text entry to coded options, in order to fully explain the health conditions of each patients [12-14]. It has also been noted that unstructured data are essential because of the information they contain [15]; therefore, unstructured data are likely to persist in the future. There is an immediate need for an automated data extraction system to transform unstructured clinical reports into a structured form, which is much easier to process and handle [16-18].

There has been considerable research in identifying patient cohorts from EHRs [19]. These approaches can be categorized into three general types: (1) rule-based approaches [20-24], (2) machine learning–based approaches [25-28], and (3) information retrieval–based approaches [29-32]. All these approaches use either pattern matching (regular expressions) or language modeling–based methods [33-36] to extract features for their system to work on. Rule-based systems are stringent and binary (either yes or no) in nature. On the other hand, machine learning– and information retrieval–based methods provide output as probability or a score. Machine learning techniques, however, require a large amount of training data to give accurate results.

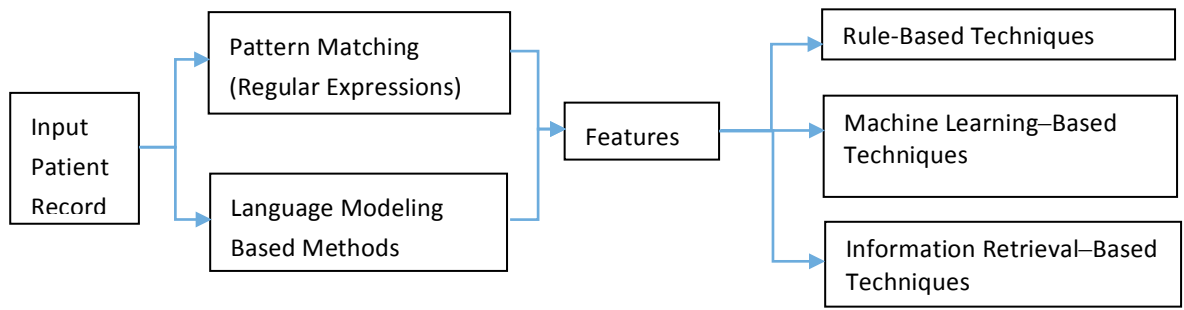

Figure 1. Summary of techniques for patient cohort identification.

Our proposed system is different from these approaches in various ways. A majority of the reported systems aim to identify whether a patient shows a certain phenotype. Therefore, the number of criteria required is less than that which are necessary for clinical trial screening. For example, a majority of the systems only use a variation of disease names, medications used, or treatments taken as their eligibility criteria [21, 23]. Ours is a more diverse application. Our goal is to check whether a particular patient qualifies for a certain clinical trial. Clinical trials usually have a large number of eligibility criteria that need to be checked. Therefore, a large amount of information related to the eligibility criteria needs to be extracted.

Our study goal is similar to that of the plethora of approaches proposed in computer-based clinical trial recruitment systems [37]. However, a majority of these approaches either lack EHRs as data source or are not equipped to handle unstructured data. We, on the other hand, obtain patient data from EHRs and handle unstructured data through information extraction methods, as opposed to the "bag of terms" or "bag of concepts" suggested in other methods [38]. The main contributions of our study are to (1) show that automated recruitment systems can only serve as prescreening tools and to (2) develop and validate a clinical trial screening system based on information extracted from EHRs. Here, we demonstrate how our system processes a set of eligibility criteria, extracts information from patient records automatically into a structured format, and finally prescreens the patients who could qualify for the trial by matching the structured patient document with the eligibility criteria.

Section 3 describes the data and the algorithm used to convert the data into a structured form. We present our results in Section 4, discuss our experience and the challenges faced in Section 5, and then conclude in Section 6.

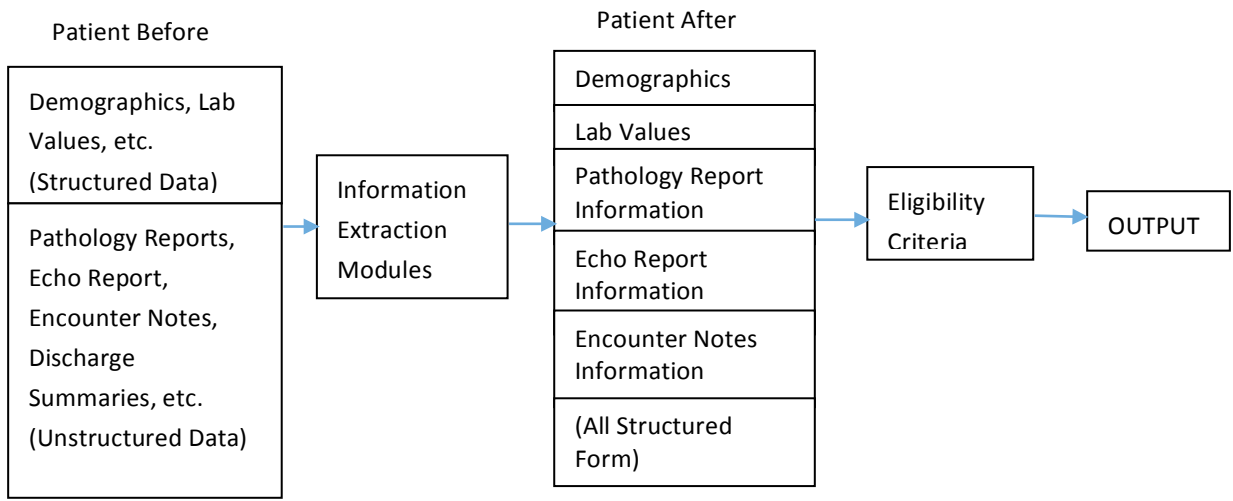

Figure 2. Overview of our clinical trial recruitment system architecture. We have analyzed different HF-related patient medical reports and derived pattern-based Information extraction modules that provide output of structured data to compare against eligibility criteria for clinical trial recruitment

## 3. Methods

### 3.1. Patient Records and Eligibility Criteria – Data Description

The patient records used in this study come from the EPIC EHR used by Northwestern Memorial Group. The initial cohort of patients we have considered for our experiments was very broad to ensure we were not missing any patients that could be included – patients that currently have been documented to have HF with the ICD-9-CM Diagnosis Code 428.0 or had an echocardiogram within the past year. We selected 198 of these patients for development and 3002 patients for validation.

Each patient's data consists of five types of reports: encounters, problem list, echocardiography reports, lab reports, and current medication list. Encounters contain two types of files: encounter diagnosis name and encounter progress notes. The characteristics of the patient records for both datasets are summarized in Table 1. We have 40 eligibility criteria – 7 for inclusion and 33 for exclusion – for the PARAGON clinical trial [39]. However, we currently evaluate our approach based on a subset of these criteria (Figure 3).

Table 1. Characteristics of the development and validation patient datasets.

|  | Development set | Validation set |
|---|---|---|
| **Total number of patients** | 198 | 3002 |
| **Encounters** | 54,173 | 393,482 |
| **Echocardiography reports** | 96,281 | 883,385 |
| **Lab reports** | 52,393 | 371,879 |
| **Current medication entries** | 4490 | 41,947 |
| **Problem lists** | 3521 | 33,089 |

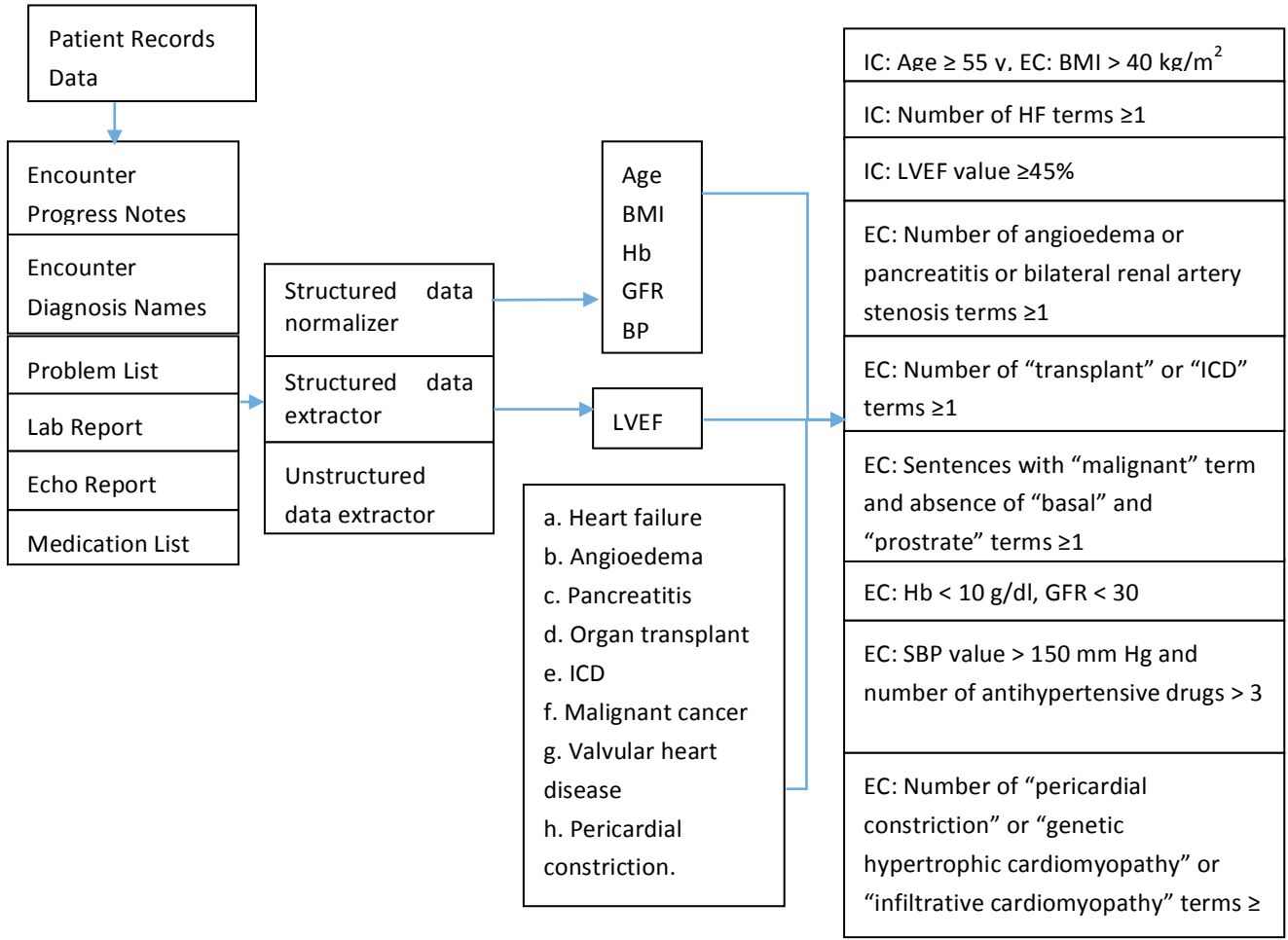

Figure 3. Algorithm. Each patient's data parsed through three types of extraction module. The modules extract the appropriate information and create a patient profile. This profile is then checked against the clinical trial eligibility criteria to check the patient's qualification. (BMI=Body Mass Index, Hb = Hemoglobin, GFR=Glomerular Filtration Rate, BP = Blood Pressure, LVEF= left ventricular ejection fraction, ICD= implantable cardioverter defibrillators, IC=Inclusion criteria, EC=Exclusion criteria)

## 3.2 Algorithm

The information from the patient data is extracted as part of separate modules. These modules are designed to extract the data elements relevant to PARAGON but are reusable individually for other clinical trials. After extraction, a rule-based system matches the eligibility criteria and discards patients who do not satisfy any of the inclusion criteria or satisfies one of the exclusion criteria. Figure 3 describes the system's architecture in finer detail. We broadly categorize the modules as: (1) structured data normalizer, (2) unstructured data extractor, and (3) unstructured data classifier.

**Structured data normalizer** is used for the extraction of data elements whose values are already present in the structured form. This module is further divided into two submodules. In submodule 1, we extract the values for age, BMI, hemoglobin, GFR and blood pressure from structured fields. In submodule 2, we extract the number of medications a patient belong to different drug classes. The reports are in structured form with mapping of the medication to the patient. This submodule requires external information resources, which we provide as databases to our system. Table 2 and Table 3 list the drug classes that we incorporated for the PARAGON clinical trial.

Table 2. Types of antihypertensive drugs

| Beta (β)-blockers | Dihydropyridines | Nondihydropyridines: class IV antiarrhythmics | Antihypertensives (others) |
|---|---|---|---|
| Acebutolol | Amlodipine | Diltiazem | Aliskiren |
| Atenolol | Felodipine | Verapamil | Fenoldopam |
| Betaxolol | Nicardipine | | Hydralazine |
| Bisoprolol | Nifedipine | | Hydralazine/HCTZ |
| Carvedilol | Nisoldipine | | Methyldopa/HCTZ |
| Esmolol | | | Minoxidil |

Table 3. Common angiotensin-converting enzyme (ACE) inhibitors and angiotensin II receptor blockers (ARBs)

| Common ACE Inhibitors | Common ARBs |
|---|---|
| Benazepril | Candesartan |
| Captopril | Eprosartan |
| Enalapril | Irbesartan |
| Fosinopril | Losartan |
| Lisinopril | Olmesartan |

**Unstructured data extractor** is used for the extraction of values of data elements present in unstructured text. This module also accepts input and provides output just as the previous module but uses a complex set of regular expressions to extract the exact value. For example, we currently extract the left ventricular ejection fraction (LVEF) value for our clinical trial. For this, we first use a set of regular expressions to extract sentences where the LVEF value may be present and then another set of regular expressions to extract the definite values as shown in Table 4. Regular expressions 1 through 4 extract the sentences that can contain LVEF values. Then, the sentences are parsed through regular expressions 5 and 6. Regular expression 5 extracts the values present in range format: for example, "40% to/- 45%." Regular expression 6 extracts the freely available values: for example, "40%."

Table 4. Regular expressions for extracting LVEF-containing sentences and values.

| S/N | Regular Expression |
|---|---|
| 1 | (left ventricular ejection fraction\|lvef\|lv ejection fraction\|left ventricle ejection fraction\|ejection fraction\|ef \|ejection fraction)[^_%\\.]*?([\\d-\\.]+)\\s*'?% |
| 2 | (left ventricular systolic function\|left ventricular function\|systolic function of the left ventricle\|lv systolic function\|left ventricular ejection fraction\|ejection fraction\|left ventricle)(normal\|normal global\|low normal\|well preserved\|severely reduced\|moderately decreased\|moderately depressed\|severely decreased\|severe\|markedly decreased\|markedly reduced\|severely globally reduced\|mildly decreased\|mildly depressed\|severely depressed) |
| 3 | (normal\|normal global\|low normal\|well preserved\|severely reduced\|moderately decreased\|moderately depressed\|severely decreased\|severe\|markedly decreased\|markedly reduced\|severely globally reduced\|mildly decreased\|mildly depressed\|severely depressed) |
| 4 | .*(moderate\|marked\|severe) (lv systolic dysfunction\|left ventricular dysfunction\|left ventricular systolic dysfunction).* |
| 5 | ((\\d+\\s*(\\-\|to)\\s*\\d+)\|(\\d*\\.\\d*\\s*(\\-\|to)\\s*\\d*\\.\\d*)\|(\\d*\\.\\d+)\|(\\d+))(?=(\\s*(\\%))) |
| 6 | \\d+(\\.\\d+)? |

**Unstructured data classifier** is used for classifying whether certain data elements are present or absent in relation to the context of the patient. Currently in this module, we extract all the instances of a given data element (diagnosis, medication, treatment, or tests) and its synonyms in the input report(s). For this, the module first checks for synonyms of the input term using UMLS Metathesaurus [40], builds automatically a set of regular expressions, and then applies them to the input report text to extract all the instances. For example, to extract HF-related terms the module compiles a list of synonyms: "heart failure," "HF," "diastolic dysfunction," and "cardiomyopathy." Next, a set of regular expressions are automatically generated (Table 5) and used to extract all the instances of HF-related terms. For PARAGON, the other data elements processed in this category are "angioedema", "pancreatitis", "valvular heart disease," etc. We adapt existing rule-based systems to make sure the data elements are not in their negated form using a rule-based negation detection algorithm, the data elements refer to the current status (as opposed to historical condition or a hypothesis for conducting a test) and the data elements correspond to the patient (as opposed to a family member or relative) [41, 42].

Table 5. Regular expressions to extract HF-related terms.

| Regular Expression |
|---|
| [^\w]+(h\|H)eart\s+(f\|F)ailure[^\w]+ |
| [^\w]+(d\|D)iastolic\s+(d\|D)ysfunction[^\w]+ |
| [^\w]+(c\|C)ardiomyopathy[^\w]+ |

```
[^\w]+HF[^\w]+
```

## 4. Evaluation and Results

We first evaluated our methods iteratively using the development set of 198 patient reports. A study coordinator read each patient record, extracted data elements of relevance to PARAGON, and matched against the eligibility criteria. For the 198 patient reports, our experienced research coordinator took two weeks (80 hours) to generate the gold standard data. Finally, we had 40 of the 198 patients (20%) prescreened for further analysis according to the eligibility criteria. After consulting a cardiologist, the number of patients finally found eligible was 12. The sheer size of the data that the clinical investigator or research coordinator has to read through is time consuming as well as tedious (Table 1).

The number of patients finally qualifying for any clinical trial is always small. This is mostly due to the large number of stringent eligibility criteria. Therefore, it becomes important for an automated system to give more importance to retrieving nearly all the qualifying patients; in other words, the recall of the system should be close to 100%. We tuned our system in order to achieve a high recall (i.e., high sensitivity) so as not to have too many false negatives (which would result in missing potentially eligible patients). On the experiments run on the development dataset, we achieved close to 95% recall with a precision of 86% (F-score of 90%). Table 6 presents further details.

Table 6. Outputs for the development dataset.

|  |  | Prescreening Gold Standard (Manual) | |
|---|---|---|---|
|  |  | Patients Included | Patients Excluded |
| Classification outcome (algorithmic) | Patients included | 38 | 6 |
|  | Patients excluded | 2 | 152 |

On the validation dataset, we prescreened 113 (3.7%) patients for the PARAGON clinical trial. Our clinical trial study coordinator went through these records and found that 21 of the patients fully qualify for the clinical trial. Twenty-five of the patients require consultation with a cardiologist. However, 67 of the patients do not qualify for the trial. In most cases, this is not because of errors in the prescreening system but due to certain other criteria that have either been not included in the algorithm (for example, certain specific allergies to medication, pregnancy, patient not present in the country, etc.) or are beyond the scope of any system to check due to lack of data (for example, type of cancer or cancer is malignant or benign when the details are not present). We detail some these issues in the Discussion section.

Table 7 lists the number of patients we discard based on each criteria for both the development and validation dataset. It can be seen that each information extraction module played a major role in screening out large proportions of patients without human involvement. For example, module 2, which extracts LVEF values, discarded 90 patients from the 198-patient development dataset and 672 patients from the 3002-patient validation dataset. This would not have been captured by any methods that aim to prioritize patients using information retrieval approaches without first extracting the values of the relevant data elements from unstructured reports.

Table 7. Number of patients discarded at each step for the 198- and 3002-patient datasets

| Criterion based on | Report Type | Number of Patients Discarded | |
|---|---|---|---|
| Age + BMI | Encounter report | 22 | 1071 |
| HF related term | Encounter report/problem list | 3 | 1597 |
| LVEF | Echo report | 90 | 672 |
| Angioedema, Pancreatitis or Bilateral Renal Artery stenosis | Encounter report | 44 | 218 |
| Organ Transplant or ICD | Encounter report/problem list | 50 | 806 |
| Malignant organ system | Encounter report/problem list | 49 | 600 |
| Hb + GFR | Lab value report | 42 | 507 |
| Blood Pressure and Hypertensive drugs | Encounter report | 6 | 522 |
| Pericardial constriction or genetic hypertrophic cardiomyopathy or infiltrative cardiomyopathy | Echo report | 23 | 314 |

The time taken by our system to successfully parse and extract the required information from different data reports is just 2 minutes (for the whole 198-patient dataset). For 3002 patients, we are able to do so in approximately 20 minutes. Our clinical trial coordinator took almost two weeks to go through each of 198 patients' reports. Thus, the time required for her to go through 3002 patients would have been several months. Instead, she only had to examine the 113 prescreened patients from our system, which only took one week. This demonstrates the usefulness of our system in practical application. However, from these results and observations, we also understand that the system can only be used for prescreening, and further validation by the clinical trial study coordinator or clinical investigator is still required.

## 5. Discussion

We achieved high recall with reasonable precision on our development dataset and were able to replicate the performance on a larger dataset. As with any automated system, there are certain limitations to our proposed architecture, which can be broadly categorized into (1) data processing and (2) data-handling issues. We briefly describe some of these issues. The precision of the system suffers from the complexity of text data. In some cases current unstructured data extractor module is unable to extract terms correctly. For example, the module fails to identify certain HF or ICD related terms. This is due to large number of synonyms and spelling mistakes for the relevant data elements.

As mentioned earlier, there are some cases where a patient has certain allergies or may show a certain adverse reaction to a medication, both of which are difficult to extract from unstructured notes because they are not always reported in a standard format within the EHR. There are also cases where the patient has moved out of the hospital's geographic area and therefore cannot provide consent for the clinical trial. These are details that are too patient-specific for automated extraction and can only be checked manually.

In some cases, the LVEF value (which is an important factor for inclusion in HF clinical trials) is present in the form of a range or qualitative description. This created a problem while checking for eligibility according to the criterion given. For example, in our clinical trial, we have set the lower limit of LVEF at 45% based on the inclusion criteria. This creates a problem when the value is contained within the range extracted (40% to 45% or 30% to 50%, for example). Our initial approach was to take the average value and compare it with the threshold. However, after consultation with the cardiologist, our approach was deemed inappropriate. Therefore, we subsequently modified our algorithm to include these patients but with a warning regarding their LVEF value. This then served as an indication to the study coordinators to recheck the echocardiogram report (and review the echocardiographic images with the clinical investigator) in order to make further decisions about the patient's eligibility for the clinical trial.

There are also some cases where the clinicians are just screening the patient for a particular diagnosis but the patient may not actually have the disease, such as a "malignancy of organ system" check of exclusion criteria. To handle this, we do not discard those patients if we find the "screening" term in the sentences extracted for eligibility check. In similar cases, we also see the term "cancer" instead of "malignancy." However, we cannot discard all patients with the "cancer" term present since some can have a benign diagnosis and not be malignant, and it is impossible for our system to decide if the cancer is malignant or relatively benign. To mitigate these issues, we currently just display a warning in these cases, as we did for LVEF. The coordinator can then perform further checks and decide the classification. In other exclusion criterion where we have to check the B-type natriuretic peptide and glomerular filtration rate values, we face the issues of non-availability and potential outliers in the data. For such cases too, we currently report them as a warning to coordinators for further checking.

We also had to deal with data-handling issues in some cases. For example, in criteria where we have to perform a check for recent hemoglobin values, we found that the value may also be present in reports other than just blood reports. To mitigate this issue, we check for hemoglobin values in all reports and then extract the most recent one. Similarly, there were also cases where "end-date" of medication and "department-name" for encounter reports were missing or misplaced. We handled such cases following discussions with the data warehouse coordinator. To summarize, we can deduce that the patient data records are noisy due to various reasons and a preprocessing module is required to handle these issues.

## 6. Conclusions and Future Work

We have presented here a new method for automated clinical trial recruitment system. We have shown, through our results and discussion, that any automated recruitment system suffices as a prescreening process that significantly reduces the workload in recruiting patients, even if it cannot completely replace manual intervention. Our system works on the hypothesis that the performance can be greatly enhanced by converting unstructured free clinical text into a structured form. To validate our hypothesis, we built modules that extract key data elements from the unstructured text on the basis of given eligibility criteria. We evaluated our system on two datasets: one of 198 patients and one of 3002 patients. Our experiments show highly favorable results and affirm our hypothesis. For future research, we aim to evaluate the reproducibility of our system for PARAGON trial at other institutions. We also intend to build further modules to use the framework for other clinical trials.

**Acknowledgments**

This work was made possible by funding from the National Library of Medicine: R00LM011389 and R01LM011416 and Novartis. Dr. Sanjiv Shah is supported by grants from the National Institutes of Health (R01 HL107577 and R01 HL127028). The authors acknowledge Prasanth Nannapaneni for his valuable ideas on extracting information from EHR.

**Disclosures:** Dr. Shah reports receiving consulting fees from Novartis, Bayer, AstraZeneca, and Alnylam.

**References**

1. Jensen, P.B., L.J. Jensen, and S. Brunak, *Mining electronic health records: towards better research applications and clinical care.* Nat Rev Genet, 2012. **13**(6): p. 395-405.
2. Sullivan, J. *Subject Recruitment and Retention: Barrier to Success*. 2004 [cited 2015 27 July]; Available from: http://www.appliedclinicaltrialsonline.com/subject-recruitment-and-retention-barriers-success.
3. Heidenreich, P.A., et al., *Forecasting the future of cardiovascular disease in the United States: a policy statement from the American Heart Association.* Circulation, 2011. **123**(8): p. 933-44.
4. Mozaffarian, D., et al., *Heart disease and stroke statistics--2015 update: a report from the American Heart Association.* Circulation, 2015. **131**(4): p. e29-322.
5. Friedman, C.P., A.K. Wong, and D. Blumenthal, *Achieving a Nationwide Learning Health System.* Science Translational Medicine, 2010. **2**(57): p. 57cm29-57cm29.
6. Friedman, C. and M. Rigby, *Conceptualising and creating a global learning health system.* Int J Med Inform, 2013. **82**(4): p. e63-71.
7. Ma, X.-J., et al., *A two-gene expression ratio predicts clinical outcome in breast cancer patients treated with tamoxifen.* Cancer Cell, 2004. **5**(6): p. 607-616.
8. Strom, B.L., et al., *Detecting pregnancy use of non-hormonal category X medications in electronic medical records*. Vol. 18. 2011. i81-i86.
9. Mathias, J.S., D. Gossett, and D.W. Baker, *Use of electronic health record data to evaluate overuse of cervical cancer screening*. Vol. 19. 2012. e96-e101.
10. De Pauw, R., et al., *Identifying prognostic factors predicting outcome in patients with chronic neck pain after multimodal treatment: A retrospective study.* Man Ther, 2015. **20**(4): p. 592-7.
11. Onofrei, M., et al., *A first step towards translating evidence into practice: heart failure in a community practice-based research network.* Inform Prim Care, 2004. **12**(3): p. 139-45.
12. Johnson, S.B., et al., *An Electronic Health Record Based on Structured Narrative.* Journal of the American Medical Informatics Association : JAMIA, 2008. **15**(1): p. 54-64.
13. Zhou, L., et al., *How many medication orders are entered through free-text in EHRs?--a study on hypoglycemic agents.* AMIA Annu Symp Proc, 2012. **2012**: p. 1079-88.
14. Zheng, K., et al., *Handling anticipated exceptions in clinical care: investigating clinician use of 'exit strategies' in an electronic health records system*. Vol. 18. 2011. 883-889.

15. Raghavan, P., et al., *How essential are unstructured clinical narratives and information fusion to clinical trial recruitment?* AMIA Summits on Translational Science Proceedings, 2014. **2014**: p. 218-223.
16. Stanfill, M.H., et al., *A systematic literature review of automated clinical coding and classification systems*. Vol. 17. 2010. 646-651.
17. Jha, A.K., *The promise of electronic records: around the corner or down the road?* JAMA, 2011. **306**(8): p. 880-1.
18. Friedman, C., T.C. Rindflesch, and M. Corn, *Natural language processing: State of the art and prospects for significant progress, a workshop sponsored by the National Library of Medicine.* Journal of Biomedical Informatics, 2013. **46**(5): p. 765-773.
19. Shivade, C., et al., *A review of approaches to identifying patient phenotype cohorts using electronic health records*. Vol. 21. 2014. 221-230.
20. Nguyen, A.N., et al., *Symbolic rule-based classification of lung cancer stages from free-text pathology reports*. Vol. 17. 2010. 440-445.
21. Mia Schmiedeskamp, P.P., et al., *Use of International Classification of Diseases, Ninth Revision, Clinical Modification Codes and Medication Use Data to Identify Nosocomial Clostridium difficile Infection •* Infection Control and Hospital Epidemiology, 2009. **30**(11): p. 1070-1076.
22. Penberthy, L., et al., *Automated matching software for clinical trials eligibility: Measuring efficiency and flexibility.* Contemporary Clinical Trials, 2010. **31**(3): p. 207-217.
23. Kho, A.N., et al., *Use of diverse electronic medical record systems to identify genetic risk for type 2 diabetes within a genome-wide association study.* J Am Med Inform Assoc, 2012. **19**(2): p. 212-8.
24. Klompas, M., et al., *Automated identification of acute hepatitis B using electronic medical record data to facilitate public health surveillance.* PLoS One, 2008. **3**(7): p. e2626.
25. Mani, S., et al., *Early prediction of the response of breast tumors to neoadjuvant chemotherapy using quantitative MRI and machine learning.* AMIA Annu Symp Proc, 2011. **2011**: p. 868-77.
26. Van den Bulcke, T., et al., *Data mining methods for classification of Medium-Chain Acyl-CoA dehydrogenase deficiency (MCADD) using non-derivatized tandem MS neonatal screening data.* J Biomed Inform, 2011. **44**(2): p. 319-25.
27. Zhao, D. and C. Weng, *Combining PubMed knowledge and EHR data to develop a weighted bayesian network for pancreatic cancer prediction.* J Biomed Inform, 2011. **44**(5): p. 859-68.
28. Kawaler, E., et al., *Learning to predict post-hospitalization VTE risk from EHR data.* AMIA Annu Symp Proc, 2012. **2012**: p. 436-45.
29. Lowe, H.J., et al., *STRIDE--An integrated standards-based translational research informatics platform.* AMIA ... Annual Symposium proceedings / AMIA Symposium. AMIA Symposium, 2009. **2009**: p. 391-395.
30. Gregg, W., et al., *StarTracker: an integrated, web-based clinical search engine.* AMIA ... Annual Symposium proceedings / AMIA Symposium. AMIA Symposium, 2003: p. 855.
31. Hanauer, D.A., et al., *Supporting information retrieval from electronic health records: A report of University of Michigan's nine-year experience in developing and using the Electronic Medical Record Search Engine (EMERSE).* Journal of Biomedical Informatics, 2015. **55**: p. 290-300.
32. Zalis, M. and M. Harris, *Advanced Search of the Electronic Medical Record: Augmenting Safety and Efficiency in Radiology.* Journal of the American College of Radiology, 2010. **7**(8): p. 625-633.
33. Lehman, L.W., et al., *Risk stratification of ICU patients using topic models inferred from unstructured progress notes.* AMIA Annu Symp Proc, 2012. **2012**: p. 505-11.
34. Carroll, R.J., A.E. Eyler, and J.C. Denny, *Naive Electronic Health Record phenotype identification for Rheumatoid arthritis.* AMIA Annu Symp Proc, 2011. **2011**: p. 189-96.
35. Liao, K.P., et al., *Electronic medical records for discovery research in rheumatoid arthritis.* Arthritis Care & Research, 2010. **62**(8): p. 1120-1127.
36. Bejan, C.A., et al., *Pneumonia identification using statistical feature selection*. Vol. 19. 2012. 817-823.
37. Kopcke, F. and H.U. Prokosch, *Employing computers for the recruitment into clinical trials: a comprehensive systematic review.* J Med Internet Res, 2014. **16**(7): p. e161.

38. Ni, Y., et al., *Automated clinical trial eligibility prescreening: increasing the efficiency of patient identification for clinical trials in the emergency department.* J Am Med Inform Assoc, 2015. **22**(1): p. 166-78.
39. *PARAGON Inclusion/Exclusion Criteria*. 2015 [cited 2015 10th August]; Available from: https://sjonnalagadda.files.wordpress.com/2015/08/paragon_ie-criteria_10-01-2014.pdf.
40. Bodenreider, O., *The Unified Medical Language System (UMLS): integrating biomedical terminology.* Nucleic Acids Res, 2004. **32**(Database issue): p. D267-70.
41. Harkema, H., et al., *ConText: an algorithm for determining negation, experiencer, and temporal status from clinical reports.* J Biomed Inform, 2009. **42**(5): p. 839-51.
42. Mitchell, K.J., et al., *Implementation and evaluation of a negation tagger in a pipeline-based system for information extract from pathology reports.* Stud Health Technol Inform, 2004. **107**(Pt 1): p. 663-7.